\documentclass[sigconf]{acmart}

\usepackage{multirow}
\usepackage{makecell}
\usepackage{subfigure}
\AtBeginDocument{%
  }

\copyrightyear{2025} 
\acmYear{2025} 
\setcopyright{acmlicensed}
\acmConference[MM '25]{Proceedings of the 33rd ACM International Conference on Multimedia}{October 27--31, 2025}{Dublin, Ireland}
\acmBooktitle{Proceedings of the 33rd ACM International Conference on Multimedia (MM '25), October 27--31, 2025, Dublin, Ireland}
\acmDOI{10.1145/3746027.3754948}
\acmISBN{979-8-4007-2035-2/2025/10}

\begin{document}

\title{Exploring Fourier Prior and Event Collaboration for Low-Light Image Enhancement}

\author{Chunyan She}
\orcid{0000-0001-8188-938X}
\affiliation{%
  \institution{College of Artificial Intelligence, Southwest University}
  \city{Chongqing}
  \country{China}
}
\email{shecyy@email.swu.edu.cn}

\author{Fujun Han}
\orcid{0009-0000-6823-4645}
\affiliation{
  \institution{College of Artificial Intelligence, Southwest University}
  \city{Chongqing}
  \country{China}
}
\email{fjhanswu@gmail.com}

\author{Chengyu Fang}
\orcid{0000-0002-6522-3710}
\affiliation{
  \institution{Shenzhen International Graduate School, Tsinghua University}
  \city{Shenzhen}
  \country{China}
}
\email{chengyufang.thu@gmail.com}

\author{Shukai Duan}
\orcid{0000-0002-0040-3796}
\affiliation{
  \institution{College of Artificial Intelligence, Southwest University}
  \city{Chongqing}
  \country{China}
}
\email{duansk@swu.edu.cn}

\author{Lidan Wang}
\orcid{0000-0003-0730-4202}
\authornote{Corresponding author}
\affiliation{
  \institution{College of Artificial Intelligence, Southwest University}
  \city{Chongqing}
  \country{China}
}
\email{ldwang@swu.edu.cn}

\renewcommand{\shortauthors}{Chunyan She, Fujun Han, Chengyu Fang, Shukai Duan, and Lidan Wang}

\begin{abstract}
The event camera, benefiting from its high dynamic range and low latency, provides performance gain for low-light image enhancement. Unlike frame-based cameras, it records intensity changes with extremely high temporal resolution, capturing sufficient structure information. Currently, existing event-based methods feed a frame and events directly into a single model without fully exploiting modality-specific advantages, which limits their performance. Therefore, by analyzing the role of each sensing modality, the enhancement pipeline is decoupled into two stages: visibility restoration and structure refinement. In the first stage, we design a visibility restoration network with amplitude-phase entanglement by rethinking the relationship between amplitude and phase components in Fourier space. In the second stage, a fusion strategy with dynamic alignment is proposed to mitigate the spatial mismatch caused by the temporal resolution discrepancy between two sensing modalities, aiming to refine the structure information of the image enhanced by the visibility restoration network. In addition, we utilize spatial-frequency interpolation to simulate negative samples with diverse illumination, noise and artifact degradations, thereby developing a contrastive loss that encourages the model to learn discriminative representations. Experiments demonstrate that the proposed method outperforms state-of-the-art models.
\end{abstract}





\maketitle

\section{Introduction}
Images captured in low-light scenarios suffer from poor visibility and blurring. These degradations are undesirable for the human visual perception system. Currently, a variety of low-light image enhancement (LLIE) methods have been proposed to improve the perceptual quality of low-light images.

\begin{figure}[!t]
  \centering
  \includegraphics[width=0.453\textwidth]{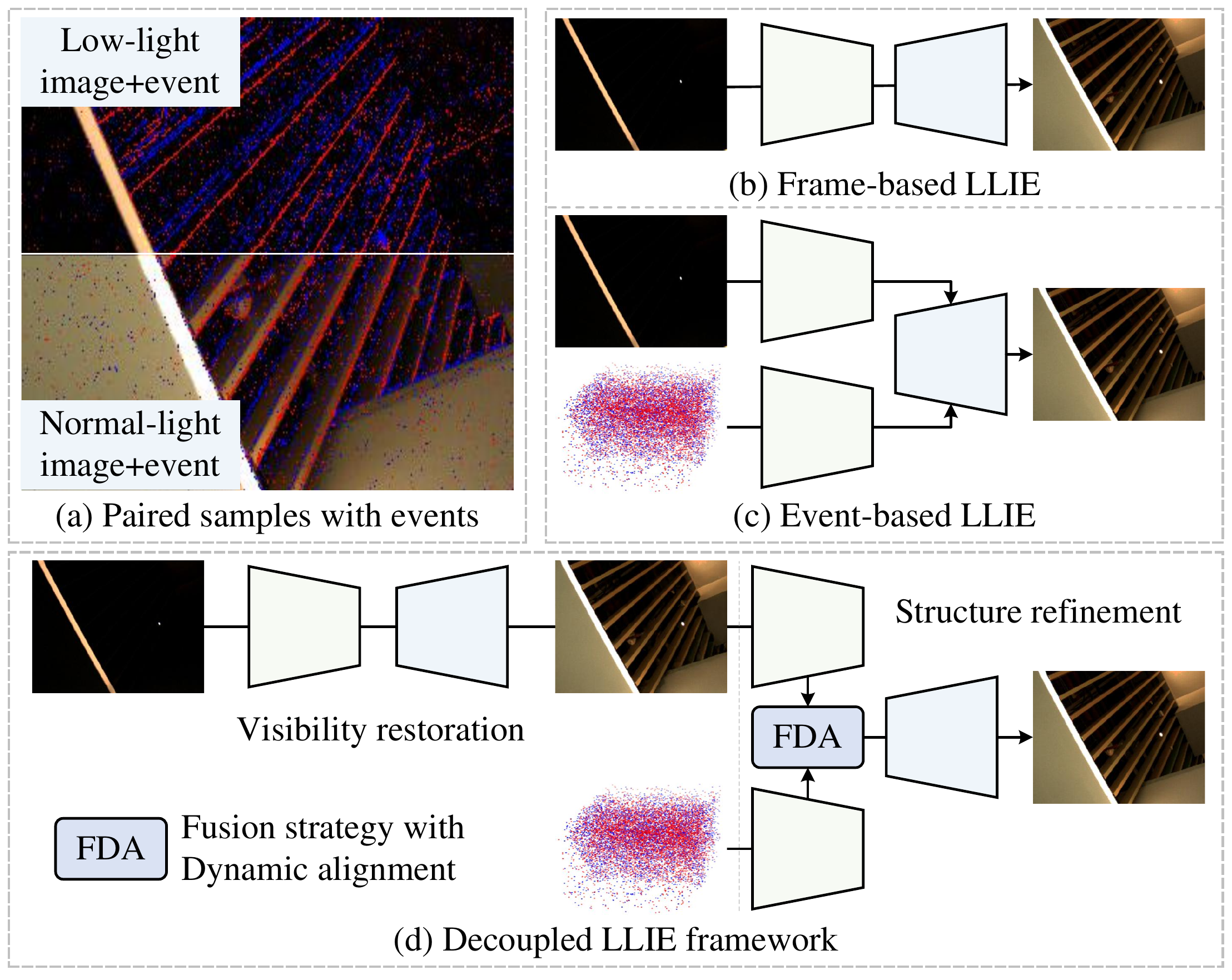}
  \caption{Paired samples with events and architectures of different LLIE paradigms. (a) shows a low-light image, a normal-light image and events. (b) The architecture of frame-based LLIE. (c) The architecture of event-based LLIE. (d) The architecture of the proposed LLIE. Compared to the existing LLIE, it is aware of the fact that the frame and events carry content and structure information, respectively. Based on this observation, the enhancement pipeline is decoupled into two stages: visibility restoration and structure refinement.}
  \label{fig1}
\end{figure}

In the past decade, frame-based LLIE \cite{bib14,bib15,bib16} has achieved promising performance, particularly with the emergence of advanced vision models \cite{bib2,bib3}. We find that the key of these methods is how to design a high-performance network to improve the quality of enhanced images. Unfortunately, it is difficult for the frame-based camera to capture complete structure details in extreme low-light scenarios, which limits the performance of frame-based LLIE.

Recently, the event camera has gradually received attention for low-level vision tasks \cite{bib4,bib5,bib6}. It is a bio-inspired vision sensor with high dynamic range and low latency. Unlike the frame-based camera, it responds to light changes by triggering asynchronous events, enabling it to perform effectively even in extremely low-light scenarios \cite{bib7}. Figure \ref{fig1}(a) shows a sample captured by a Dynamic and Active-Pixel Vision Sensor (DAVIS) camera, one of the common event cameras. Obviously, the event camera can alleviate the limitation of the frame-based camera in capturing structure details in extremely low-light regions. Moreover, we can observe that the low-light image and the events carry content and structure, respectively, i.e., the events aim to provide rich structure information for LLIE, independent of visibility restoration. As shown in Figures \ref{fig1}(b) and (c), existing methods feed a low-light image and events directly into the LLIE model to reconstruct an enhanced image, i.e., visibility restoration and structure refinement are coupled in a single model without fully exploiting modality-specific advantages.

Based on the above discussions, we propose a two-stage decoupling framework called \textbf{EventLLIE}, as shown in Figure \ref{fig1}(d). Compared to the existing LLIE, it is aware of the fact that the frame and events carry content and structure information, respectively. Specifically, the enhancement pipeline is decoupled into two stages consisting of \textbf{visibility restoration} and \textbf{structure refinement}. In the first stage, inspired by the existing Fourier-based methods \cite{bib18,bib39}, we embed Fourier prior into a deep model to restore the visibility of low-light images in Fourier space. Unlike the existing methods, we rethink the relationship between amplitude and phase components, and design a visibility restoration network with amplitude-phase entanglement. In the second stage, events are utilized to refine the structure of the image enhanced by visibility restoration network. While events can adequately represent structure information, we find that there is a spatial mismatch between events and image's structure. A fusion strategy with dynamic alignment is therefore designed to integrate multi-modal representations in structure refinement network. Furthermore, we utilize spatial-frequency interpolation to generate negative samples exhibiting illumination variations, noise patterns, and artifacts, aiming to construct a contrastive loss to eliminate undesirable degradations in enhanced images. In general, our contributions are as follows:
\begin{itemize}
\item By analyzing the roles of images and events in LLIE task, a two-stage LLIE model is built to decouple the enhancement pipeline into two stages consisting of visibility restoration and structure refinement.
\item In the first stage, we design a visibility restoration network with amplitude-phase entanglement by rethinking the relationship between amplitude and phase components.
\item In the second stage, a fusion strategy with dynamic alignment driven by the similarity between image content and event information is proposed to organically fuse them, aiming to refine the structure of enhanced images.
\item We develop a spatial-frequency interpolation-based contrastive loss in which the enhanced image is forcibly pushed away from the semantic space with diverse degradations.
\end{itemize}

\section{Related Work}


\textbf{Traditional and Learning-based LLIE}. Frame-based LLIE can be divided into traditional and learning-based LLIE. Traditional LLIE include gamma correction \cite{bib8}, histogram equalization \cite{bib9}, Retinex \cite{bib10}, etc. They rely on artificial parameters and priors, resulting in a lack of generalization. With the emergence of deep learning, learning-based LLIE has achieved competitive performance. Notably, a LLIE based on convolutional neural network was first proposed in \cite{bib11}. Wei et al. \cite{bib13} and Cai et al. \cite{bib14} developed deep models based on Retinex theory. In \cite{bib15}, Xu et al. designed a transformer-based enhancer that adopted a Signal-to-Noise Ratio (SNR) map to avoid the interference of noise. Considering the importance of structure information, She et al. \cite{bib12,bib44} employed a high-level vision model to extract structure priors, which were then leveraged to encourage the enhancer to learn representations with rich structure. \cite{bib16} utilized neural implicit representation to enhance low-light images in Hue, Saturation and Value (HSV) color space. While the HSV color space decouples the illumination channel, it tends to introduce noise and artifacts. To this end, Yan et al. \cite{bib41} proposed a color space called Horizontal/Vertical-Intensity. Based on this novel color space, a decoupling model was developed to learn illumination mapping under different illumination conditions. Given the success of the diffusion model \cite{bib32,bib43}, He et al. \cite{bib32} utilized the diffusion model to generate reflectance and illumination priors to guide LLIE task. Although the above methods have made promising progress, they directly build deep models in spatial domain to enhance the visibility of low-light images without sufficiently considering the intrinsic properties of illumination degradation.
\begin{figure*}[!t]
  \centering
  \includegraphics[width=0.945\textwidth]{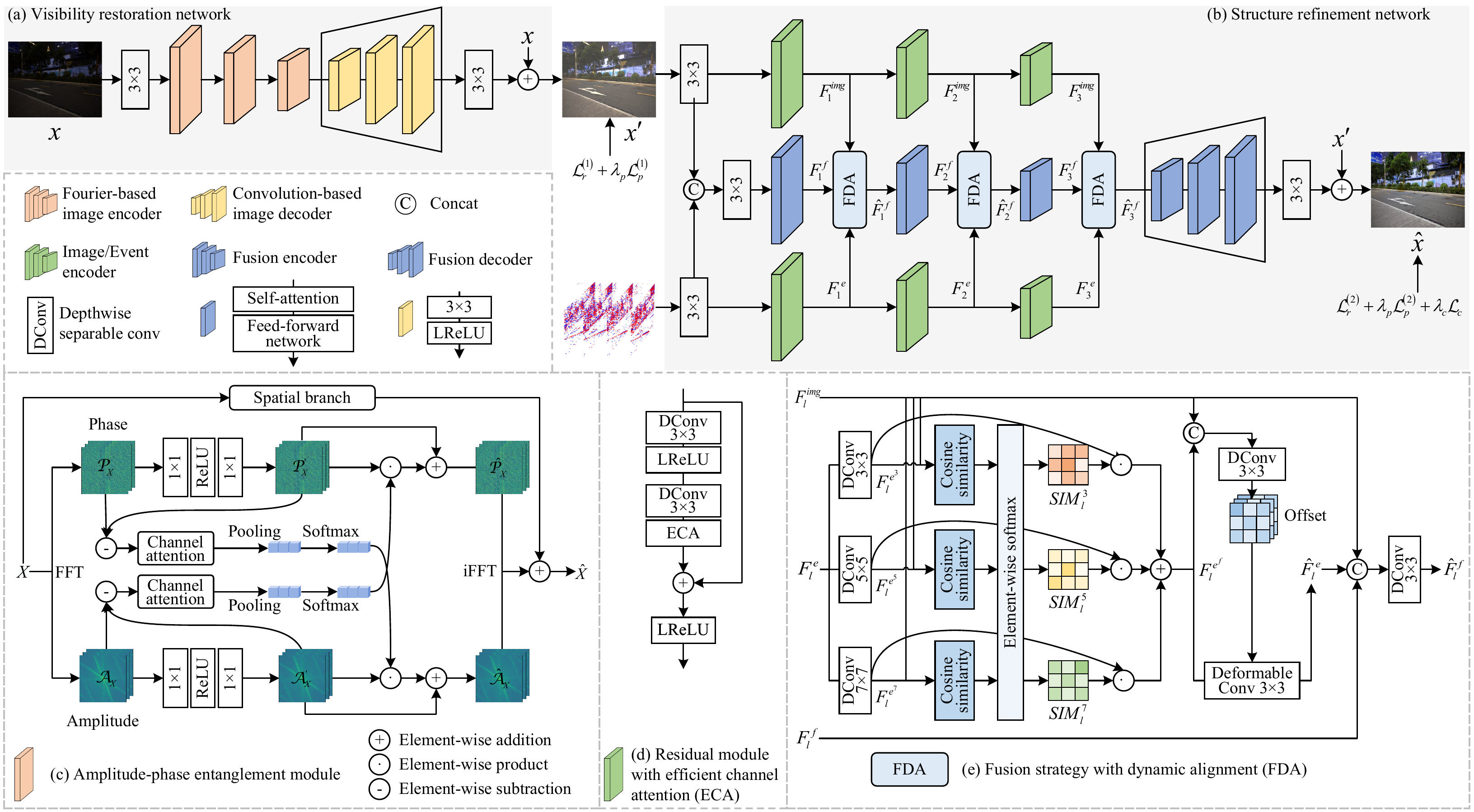}
  \caption{Overall framework of the proposed EventLLIE. A low-light image is first fed into (a) the visibility restoration network for initial enhancement, and then (b) the structure refinement network is employed to further refine the structure information. (c) The proposed amplitude-phase entanglement module. (d) Residual module with efficient channel attention \cite{bib31}. (e) The proposed fusion module with dynamic alignment. For simplicity, we omit skip connections between the Fourier-based image encoder and convolution-based decoder, and also between the fusion encoder and fusion decoder.}
  \label{fig2}
\end{figure*}

\textbf{Fourier-based LLIE}. Recently, some researchers \cite{bib17,bib18,bib19,bib20} revealed that degradation of low-light images is concentrated in the amplitude component of Fourier space. To this end, Huang et al. \cite{bib17} proposed an exposure correction network based on deep Fourier. To improve the efficiency, \cite{bib18} designed an ultra-high-definition enhancer, which restored the amplitude component in the low-resolution space. Wang et al. \cite{bib19} presented a two-stage LLIE. In the first stage, it is devoted to learning an efficient transform to enhance the amplitude component. In the second stage, an SNR map is employed to guide the reconstruction of spatial details. In \cite{bib20}, the Fourier prior was integrated into the diffusion model to guide diffusion sampling. It first forces the amplitude component to approximate the distribution of normal-light images, and then the phase component is refined to provide clear structure guidance. We can find that these methods lack attention to the phase component. Although a few works consider both amplitude and phase components, their relationship has not been explored. Unlike them, we rethink their relationship and propose a visibility restoration network with amplitude-phase entanglement.

\textbf{Event-based LLIE}. Benefiting from its high dynamic range characteristic, the event camera can capture rich structure information in extreme low-light scenarios \cite{bib42}. \cite{bib6} constructed the first real event-based dataset for LLIE and developed an enhancement method based on event streams and low-light frames. Liu et al. \cite{bib4} synthesized events through low-light video and then utilized the synthesized events to mitigate artifacts. Kim et al. \cite{bib21} proposed an end-to-end framework for joint LLIE and deblurring, which employed an event-guided feature alignment and a cross-modal feature enhancement to fuse the two types of features. Unlike them, Liang et al. \cite{bib5} did not directly fuse event and image features, but employed an SNR map to adaptively fuse them, considering that low-light images and events may be corrupted by different patterns of noise. Existing event-based methods realize that the efficacy of events is to supplement structure information, and work on designing reasonable strategies to fuse image and event features. However, two complementary modalities are coupled in a single model without fully exploiting modality-specific advantages, which limits their performance.

\section{Method}
The architecture of the proposed EventLLIE is shown in Figure \ref{fig2}, which consists of visibility restoration and structure refinement networks. Considering the promising performance of the Unet-like structure in the image restoration \cite{bib1}, it is used as the basic architecture. In the first stage, we design a visibility restoration network (see Figure \ref{fig2}(a)) with amplitude-phase entanglement to obtain coarse enhancement results. In the second phase, a structure refinement network (see Figure \ref{fig2}(b)) is employed to organically integrate events, aiming to refine the structure of enhanced results.

\subsection{Visibility Restoration}
\label{Visibility-Restoration}

\begin{figure*}[!t]
  \centering
  \includegraphics[width=0.945\textwidth]{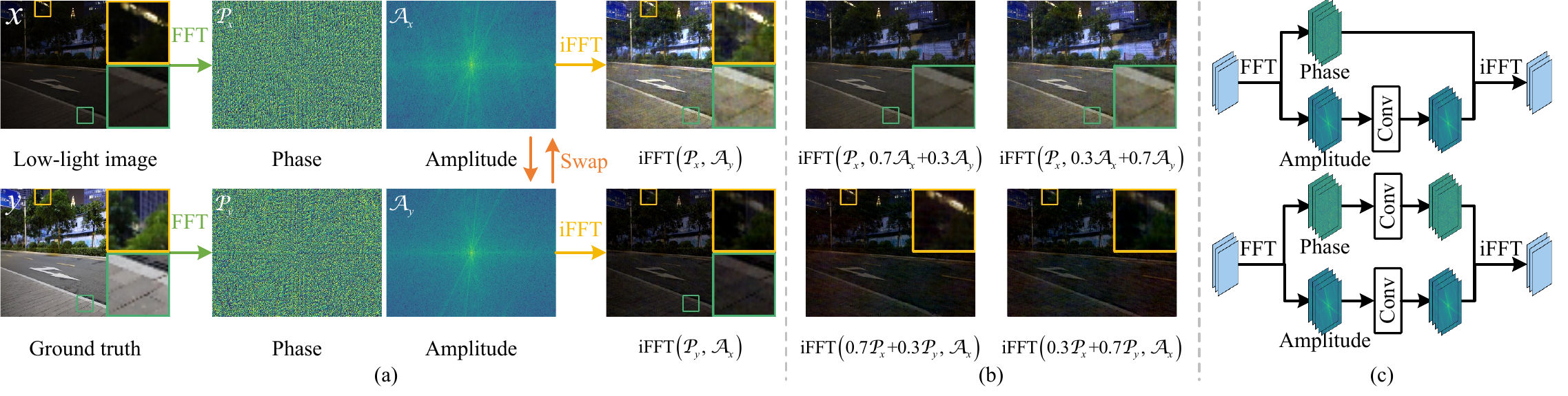}
  \caption{The analysis for Fourier-based LLIE. (a) Existing Fourier-based works \cite{bib18,bib19} reveal that the visibility of a low-light image $x$ can be restored by exchanging the amplitude components between the low-light image $x$ and the ground-truth $y$. (b) We present several reconstructed results through inverse fast Fourier transform for cases of phase interpolation and amplitude interpolation. (c) Two modules are commonly used in existing Fourier-based methods \cite{bib18,bib19}.}
  \label{fig3}
\end{figure*}

Fourier-based LLIE methods \cite{bib18,bib19} reveal that amplitude and phase components represent illumination and structure information, respectively. As shown in Figure \ref{fig3}(a), given a low-light image $x$ and a ground truth $y$, we first employ the fast Fourier transform ${\rm{FFT}}\left(  \cdot  \right)$ to obtain their amplitude components ${{\mathcal A}_{\left(  \cdot  \right)}}$ and phase components ${{\mathcal P}_{\left(  \cdot  \right)}}$ respectively, and then exchange their amplitude components. Finally, the inverse fast Fourier transform ${\rm{iFFT}}\left( { \cdot {\rm{, }} \cdot } \right)$ is used to reconstruct image ${\rm{iFFT}}\left( {{{\mathcal P}_x},{{\mathcal A}_y}} \right)$ and image ${\rm{iFFT}}\left( {{{\mathcal P}_y},{{\mathcal A}_x}} \right)$. We can observe that the visibility of the low-light image can be restored by exchanging the amplitude components between them. Therefore, existing Fourier-based methods aim at adjusting the amplitude component of the low-light image to obtain high-quality enhanced images. Unfortunately, there are undesirable artifacts in the reconstructed image ${\rm{iFFT}}\left( {{{\mathcal P}_x},{{\mathcal A}_y}} \right)$, which indicates that improving the amplitude component alone is difficult to obtain satisfactory results, i.e., the amplitude and phase components are not independent. As shown in Figure \ref{fig3}(b), we present the reconstruction results of amplitude interpolation and phase interpolation to further reveal the relationship between them. For the low-light image $x$, as the contribution of the amplitude component ${{\mathcal A}_y}$  increases, the visibility of reconstructed image ${\rm{iFFT}}\left( {{{\mathcal P}_x}{\rm{, }} \cdot } \right)$ improves, but artifacts become more severe. Similarly, the lighting visibility of reconstructed image ${\rm{iFFT}}\left( { \cdot ,{{\mathcal A}_x}} \right)$ will be disturbed as the contribution of the phase component ${{\mathcal P}_y}$  increases. Based on the above discussions, we can observe two phenomena: \textbf{1)} It is difficult to obtain satisfactory enhanced images by optimizing only the amplitude component. \textbf{2)} The variations in phase and amplitude exhibit mutual interference. Figure \ref{fig3}(c) presents two classical Fourier modules employed by existing Fourier-based works \cite{bib18,bib19}. However, they either optimize only the amplitude component or optimize the amplitude and phase components independently without considering their relationship.

\textbf{Fourier-based Image Encoder}. Based on the above observations, we propose a visibility restoration network with amplitude-phase entanglement to improve the illumination conditions of low-light images. Figures \ref{fig2}(a) and (c) show the overall architecture and the amplitude-phase entanglement module. Specifically, given a feature map $X$, we first adopt the fast Fourier transform ${\rm{FFT}}\left(  \cdot  \right)$ to obtain amplitude ${{\mathcal A}_X}$ and phase ${{\mathcal P}_X}$. Then, two convolution layers are employed to generate the primary optimized amplitude ${\mathcal A}_X^{'}$ and phase ${\mathcal P}_X^{'}$. Next, the difference features (i.e., ${{\mathcal A}_X^{'} - {{\mathcal A}_X}}$ and ${{\mathcal P}_X^{'} - {{\mathcal P}_X}}$) are fed into global average pooling and softmax function to compute two weight vectors, aiming to obtain the re-optimized amplitude ${\hat {\mathcal A}_X}$ and phase ${\hat {\mathcal P}_X}$. Here, we insert a channel attention (CA) module \cite{bib23} to strengthen the representation ability of difference features. Finally, the inverse fast Fourier transform ${\rm{iFFT}}\left( { \cdot {\rm{, }} \cdot } \right)$ is utilized to reconstruct the feature map $\hat X$. The above processing can be formally described as:
\begin{equation}
	\label{eq1}
	\begin{aligned}
    \hat X &= {\mathcal S}\left(  X  \right) + {\rm{iFFT}}\left( {{{\hat {\mathcal A}}_X},{{\hat {\mathcal P}}_X}} \right),\\
    {{\hat {\mathcal A}}_X} &= {\mathcal A}_X^{'} + {\mathcal A}_X^{'} \cdot {\mathop{\rm softmax}\nolimits} \left( {{\rm{pool}}\left( {{\rm{CA}}\left( {{\mathcal P}_X^{'} - {{\mathcal P}_X}} \right)} \right)} \right),\\
    {{\hat {\mathcal P}}_X} &= {\mathcal P}_X^{'} + {\mathcal P}_X^{'} \cdot {\mathop{\rm softmax}\nolimits} \left( {{\rm{pool}}\left( {{\rm{CA}}\left( {{\mathcal A}_X^{'} - {{\mathcal A}_X}} \right)} \right)} \right),\\
    {\mathcal A}_X^{'} &= W_2^{\mathcal A}\left( {{\mathop{\rm ReLU}\nolimits} \left( {W_1^{\mathcal A}\left( {{{\mathcal A}_X}} \right)} \right)} \right),\\
    {\mathcal P}_X^{'} &= W_2^{\mathcal P}\left( {{\mathop{\rm ReLU}\nolimits} \left( {W_1^{\mathcal P}\left( {{{\mathcal P}_X}} \right)} \right)} \right),
	\end{aligned}
\end{equation}
where ${{\mathcal A}_X},{{\mathcal P}_X} = {\mathop{\rm FFT}\nolimits} \left( X \right)$. $W\left(  \cdot  \right)$, ${\mathop{\rm ReLU}\nolimits} \left(  \cdot  \right)$, ${\rm{pool}}\left(  \cdot  \right)$ and ${\rm{CA}}\left(  \cdot  \right)$ represent convolution operation, ReLU function, global average pooling and CA module \cite{bib23}. Inspired by the Fourier-based works \cite{bib18,bib19}, a spatial branch ${\mathcal S}\left(  \cdot  \right)$ \cite{bib18} is added to the proposed amplitude-phase entanglement module to enhance its local information.

\textbf{Convolution-based Image Decoder}.
The purpose of the visibility restoration network is to employ the Fourier prior to coarsely restore the visibility of low-light images. As shown in Figure \ref{fig2}(a), to avoid over-increasing the computational burden, several $3 \times 3$ convolution with LeakyReLU are employed to build its decoder. Finally, we adopt a $3 \times 3$ convolution to generate the preliminary enhancement results. For simplicity, we omit the skip connection between the Fourier-based image encoder and the convolution-based image decoder in Figure \ref{fig2}(a).

\subsection{Structure Refinement}
\label{Structure_Refinement}
In the first stage, the visibility restoration network is utilized to improve the illuminance conditions of low-light images. In the second stage, we employ events to further refine the enhanced results from the first stage. Since the event streams cannot be fed directly into the depth model, we adopt the coding scheme proposed by Zhu et al. \cite{bib33} to transform the event streams $\left\{ {{e_i}} \right\}_{i = 1}^N$ into a 3D spatial-temporal event voxel $E$. In this work, the temporal bin of the event voxel is set to $32$. As shown in Figure \ref{fig2}(b), we first adopt an image encoder and an event encoder to extract content and structure features, respectively. A fusion encoder is then employed to fuse two complementary modalities, aiming to learn content representations with intact structure. Finally, a fusion decoder utilizes these features to produce high-quality enhanced results.

\textbf{Image and Event Encoders}. Given the success of residual learning \cite{bib30} and attention mechanism in image restoration, a simple residual module (see Figure \ref{fig2}(d)) with efficient channel attention (ECA) \cite{bib31} is built to extract image feature $F_l^{img}\left( {l = 1,2,3} \right)$ and event feature $F_l^e$. In addition, we replace the standard convolution with a depthwise separable convolution (DConv) to reduce the computational burden.

\begin{figure}[!t]
  \centering
  \includegraphics[width=0.39\textwidth]{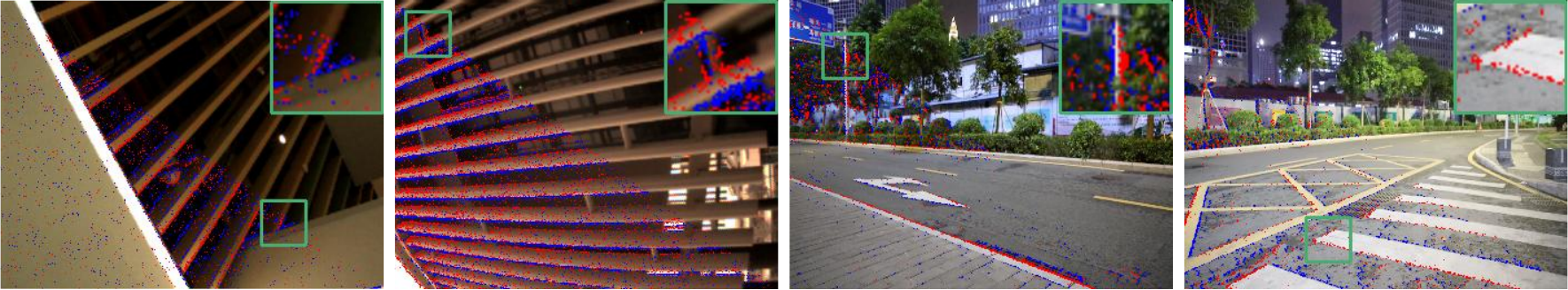}
  \caption{We show several samples with events. It can be observed that there is a spatial mismatch issue between events and image's structure, i.e., the spatial position between events and the image's structure is not exactly matched.}
  \label{fig4}
\end{figure}

\textbf{Fusion Encoder and Fusion Decoder}. Figure \ref{fig2}(b) presents the overall architecture of the fusion encoder. It first extracts a simple fusion representation by a $3 \times 3$ convolution and a concatenation operation. Then a transformer module \cite{bib29} with self-attention and feed-forward network is employed to model long-range dependence. Finally, the proposed fusion strategy is utilized to dynamically fuse image and event features. As shown in Figure \ref{fig4}, despite events can convey structure information, they do not precisely match the image's structure in spatial position. This is due to the temporal resolution discrepancy between events and frames captured by the DAVIS camera in motion scenarios. To alleviate the above issue, we propose a fusion strategy with dynamic alignment, driven by the similarity between image and event features, to obtain aligned event features. Figure \ref{fig2}(e) shows its architecture. Specifically, given the image feature $F_l^{img}$, event feature $F_l^{e}$, and fusion feature $F_l^{f}$, we first extract the multi-scale event representation $F_l^{{e^s}}\left( {s = 3,5,7} \right)$ by DConv with three kernel sizes, and then calculate the element-wise cosine similarity between multi-scale event feature $F_l^{{e^s}}$ and image feature $F_l^{img}$. Let $f_l^{e_{ij}^s}$ and $f_l^{im{g_{ij}}}$ denote the feature vectors of $F_l^{{e^s}}$ and $F_l^{img}$ at the spatial position ($i$, $j$), respectively, and the cosine similarity between them can be described as:
\begin{equation}
\label{eq2}
  {\rm{sim}}\left( {f_l^{im{g_{ij}}},f_l^{e_{ij}^s}} \right) = \frac{{{{\left( {f_l^{im{g_{ij}}}} \right)}^T}f_l^{e_{ij}^s}}}{{{{\left\| {f_l^{im{g_{ij}}}} \right\|}_2}{{\left\| {f_l^{e_{ij}^s}} \right\|}_2}}}.
\end{equation}
To dynamically adjust the contribution of multi-scale event features, we apply the softmax function to normalize their similarity maps at the same spatial position, which can be expressed as:
\begin{equation}
\label{eq3}
{\rm{si}}{{\rm{m}}^{'}}\left( {f_l^{im{g_{ij}}},f_l^{e_{ij}^s}} \right) = \frac{{\exp \left( {{\rm{sim}}\left( {f_l^{im{g_{ij}}},f_l^{e_{ij}^s}} \right)} \right)}}{{\sum\nolimits_s {\exp \left( {{\rm{sim}}\left( {f_l^{im{g_{ij}}},f_l^{e_{ij}^s}} \right)} \right)} }}.
\end{equation}
We then employ the normalized similarity map $SIM_l^s$ to obtain the aligned event feature $F_l^{{e^f}} = \sum\nolimits_s {SIM_l^s \odot F_l^{{e^s}}} $, where each element of $SIM_l^s$ represents the normalized similarity between feature $F_l^{{e^s}}$ and feature $F_l^{img}$ at each spatial position. Next, a deformable convolution is utilized to further align two-modal features. Finally, the fused feature $\hat F_l^f$ is generated by a concatenation operation and a depthwise separable convolution. Compared with image feature $F_l^{img}$ and event feature $F_l^e$, the fused feature $\hat F_l^f$ not only exhibits content representation but also encompasses rich structure details. For the fusion decoder, it consists of several transformer modules and takes these fused features to reconstruct the enhanced image. For simplicity, we omit the skip connections between fusion encoder and fusion decoder in Figure \ref{fig2}(b).

\begin{figure}[!t]
  \centering
  \includegraphics[width=0.39\textwidth]{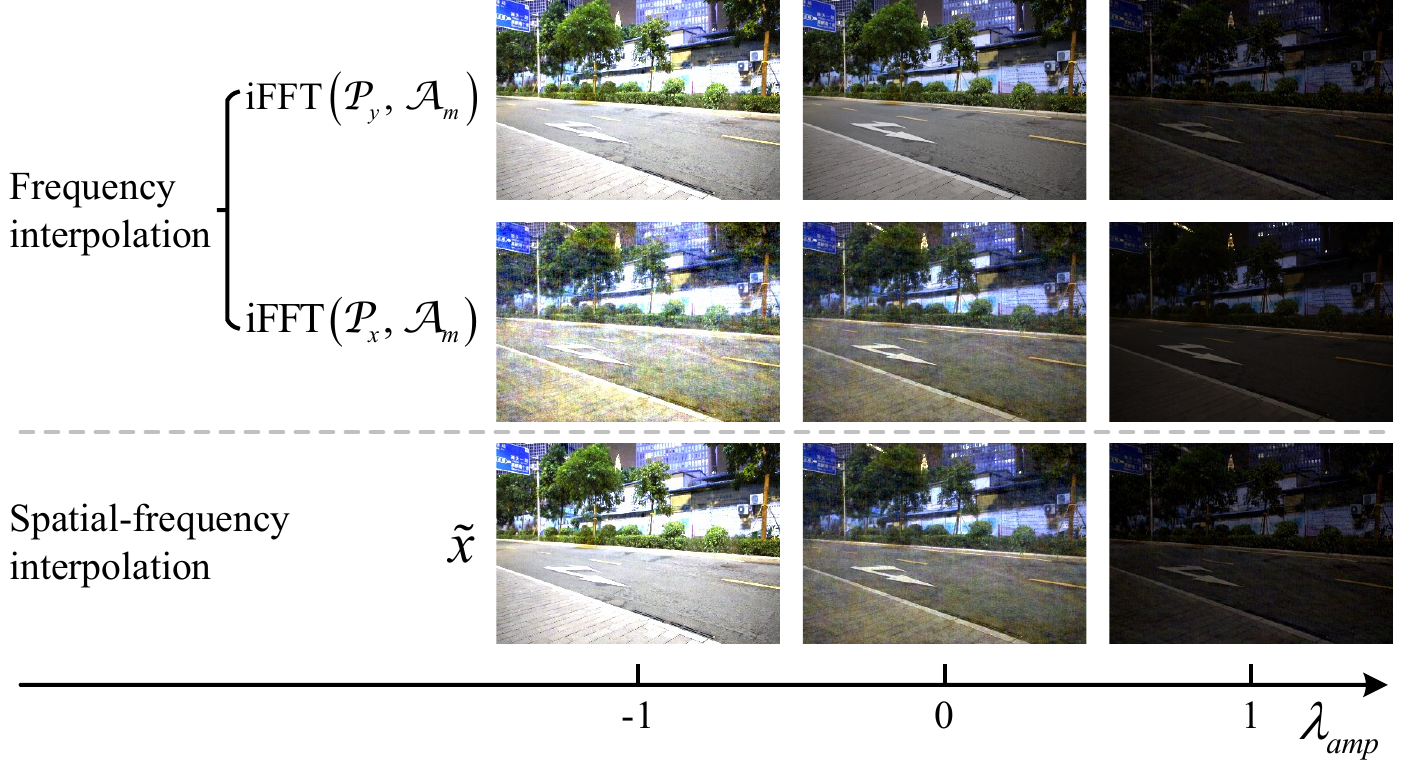}
  \caption{Samples generated by frequency and spatial-frequency interpolation. As the parameter ${\lambda _{amp}}$ tends toward 0, the frequency-domain interpolated image ${\rm{iFFT}}\left( {{{\mathcal P}_y}{\rm{, }}{{\mathcal A}_m}} \right)$ becomes highly similar to the ground truth. As the parameter ${\lambda _{amp}}$ approaches 1, the frequency-domain interpolated image ${\rm{iFFT}}\left( {{{\mathcal P}_x}{\rm{, }}{{\mathcal A}_m}} \right)$ lacks noise. We also observe that the interpolated images ${\rm{iFFT}}\left( {{{\mathcal P}_y}{\rm{, }}{{\mathcal A}_m}} \right)$ and ${\rm{iFFT}}\left( {{{\mathcal P}_x}{\rm{, }}{{\mathcal A}_m}} \right)$ enjoy complex noise in low-light and overexposure scenarios, respectively. Therefore, they are mixed again in the spatial domain to generate the negative sample $\tilde x$ with diverse degradations.}
  \label{fig5}
\end{figure}

\subsection{Training Objectives}

\begin{figure*}[!t]
  \centering
  \includegraphics[width=0.802\textwidth]{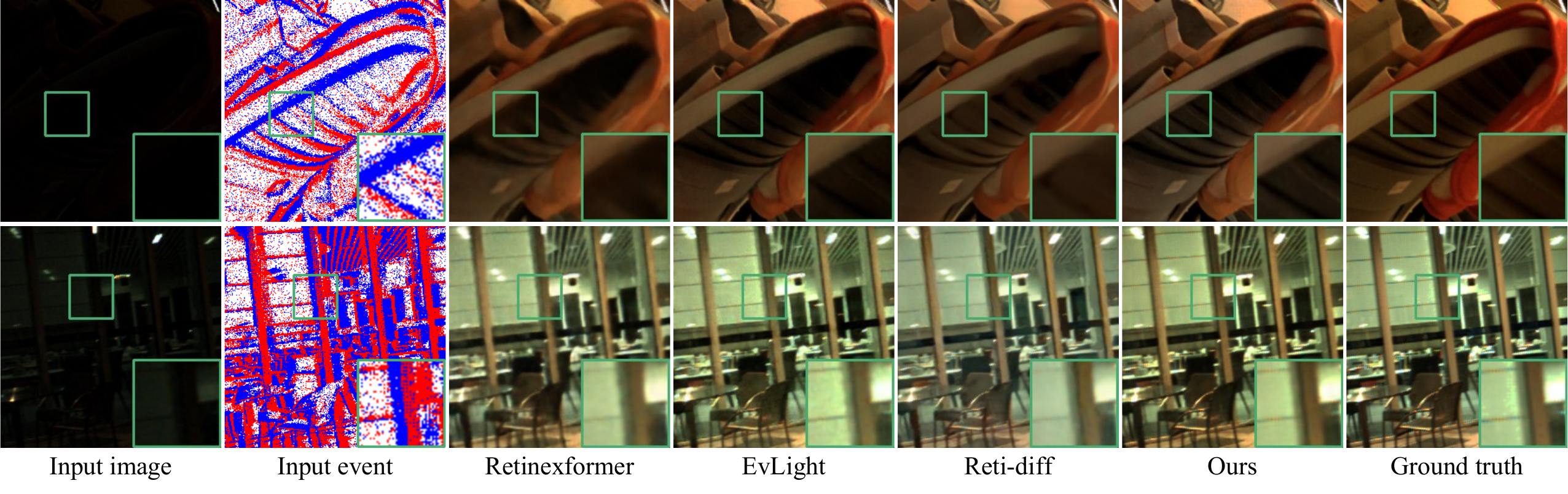}
  \caption{Visual comparison of different methods on SDE-in (the first row) and SDE-out (the second row) datasets. More enhanced images are shown in Appendix A.}
  \label{fig6}
\end{figure*}

In this work, we adopt reconstruction loss ${{\mathcal L}_{r}}$ \cite{bib24}, perception loss ${{\mathcal L}_{p}}$ \cite{bib25} to optimize the visibility restoration network. For the structure refinement network, we also employ the proposed contrastive loss ${{\mathcal L}_{c}}$ to optimize it. The overall loss function is expressed as:
\begin{equation}
\label{eq4}
{{\mathcal L}} = \underbrace {0.1 * \left( {{\mathcal L}_{r}^{\left( 1 \right)} + {\lambda _{p}}{\mathcal L}_{p}^{\left( 1 \right)}} \right)}_{{\text{visibility restoration}}} + \underbrace {\left( {{\mathcal L}_{r}^{\left( 2 \right)} + {\lambda _{p}}{\mathcal L}_{p}^{\left( 2 \right)} + {\lambda _{c}}{{\mathcal L}_{c}}} \right)}_{{\text{structure refinement}}},
\end{equation}
where ${\mathcal L}_r^{\left( 1 \right)} = \sqrt {{{\left\| {x{'} - y} \right\|}_2} + \varepsilon } $, ${\mathcal L}_{p}^{\left( 1 \right)} = {\left\| {{\mathcal F}\left( {x{'}} \right) - {\mathcal F}\left( y \right)} \right\|_1}$, ${\mathcal L}_r^{\left( 2 \right)} = \sqrt {{{\left\| {\hat x - y} \right\|}_2} + \varepsilon } $ and ${\mathcal L}_{p}^{\left( 2 \right)} = {\left\| {{\mathcal F}\left( {\hat x} \right) - {\mathcal F}\left( y \right)} \right\|_1}$. ${x{'}}$ and ${\hat x}$ represent the outputs of the visibility restoration and structure refinement networks, respectively. $y$ is the ground truth. The parameters ${\lambda _{p}}$ and ${\lambda _{c}}$ are balancing factors. The parameter $\varepsilon $ is set to 1e-6. ${\mathcal F}\left(  \cdot  \right)$ denotes the process of extracting features from the pre-trained VGG-19 network.

To further eliminate the undesirable degradations of enhanced images, we propose a contrastive loss based on spatial-frequency interpolation. The key to contrastive loss is to pull enhanced images towards positive samples and push away from negative samples. Therefore, it is natural and reasonable to consider low-light image and ground truth as negative and positive samples, respectively \cite{bib26}. Li et al. \cite{bib34} reveal that increasing the diversity of negative samples can improve the performance of contrastive loss. To this end, we utilize spatial-frequency interpolation to generate negative samples with diverse degradations. Let ${{\mathcal A}_x}$ and ${{\mathcal P}_x}$ denote the amplitude and phase components of the low-light image $x$, and ${{\mathcal A}_y}$ and ${{\mathcal P}_y}$ denote the amplitude and phase components of the ground truth $y$. We first obtain the frequency-domain interpolated amplitude component ${{\mathcal A}_m}$ between ${{\mathcal A}_x}$ and ${{\mathcal A}_y}$ as:
\begin{equation}
\label{eq5}
{{\mathcal A}_m} = {\lambda _{amp}}{{\mathcal A}_x} + \left( {1 - {\lambda _{amp}}} \right){{\mathcal A}_y},
\end{equation}
where ${\lambda _{amp}} \sim U\left( { - 1,1} \right)$. The interpolated amplitude ${{\mathcal A}_m}$ is then employed to reconstruct two interpolated images ${\rm{iFFT}}\left( {{{\mathcal P}_x}{\rm{, }}{{\mathcal A}_m}} \right)$ and ${\rm{iFFT}}\left( {{{\mathcal P}_y}{\rm{, }}{{\mathcal A}_m}} \right)$. As shown in Figure \ref{fig5}, to enrich the diversity of negative samples, the parameter ${\lambda _{amp}}$ will be sampled from a uniform distribution $U\left( { - 1,1} \right)$, which encourages interpolated images to simulate underexposed and overexposed samples. However, when parameter ${\lambda _{amp}}$ is approximately equal to $0$, the interpolated image ${\rm{iFFT}}\left( {{{\mathcal P}_y}{\rm{, }}{{\mathcal A}_m}} \right)$ will be highly similar to the ground truth $y$, which will cause it to fail as a negative sample. To avoid this issue and further increase the diversity of negative samples, two frequency-domain interpolated images are mixed again in the spatial domain to generate the negative sample $\tilde x$. The spatial interpolation can be formulated as:
\begin{equation}
\label{eq6}
\tilde x = \left( {1 - \left| {{\lambda _{amp}}} \right|} \right){\rm{iFFT}}\left( {{{\mathcal P}_x}{\rm{, }}{{\mathcal A}_m}} \right) + \left| {{\lambda _{amp}}} \right|{\rm{iFFT}}\left( {{{\mathcal P}_y}{\rm{, }}{{\mathcal A}_m}} \right).
\end{equation}

\begin{figure*}[!t]
  \centering
  \includegraphics[width=0.802\textwidth]{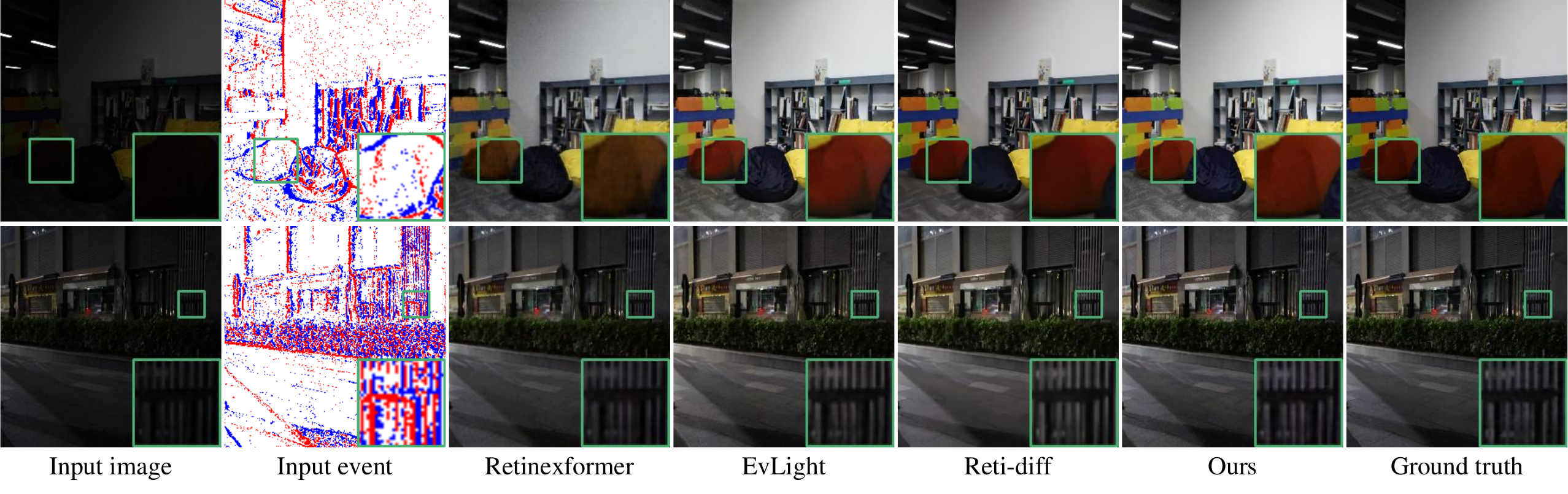}
  \caption{Visual comparison of different methods on SDSD-in (the first row) and SDSD-out (the second row) datasets. More enhanced images are shown in Appendix A.}
  \label{fig7}
\end{figure*}

With the generation of negative samples, the proposed contrastive loss is defined as:
\begin{equation}
\label{eq7}
{{\mathcal L}_{c}} = \frac{{{{\left\| {{\mathcal F}\left( {\hat x} \right) - {\mathcal F}\left( y \right)} \right\|}_1}}}{{\frac{1}{K}\sum\limits_{k = 1}^K {{{\left\| {{\mathcal F}\left( {\hat x} \right) - {\mathcal F}\left( {{{\tilde x}_k}} \right)} \right\|}_1}} }},
\end{equation}
where ${\tilde x_k}$ represents the $k$th negative sample generated by Eq. (\ref{eq6}), and parameter $K$ is the number of negative samples. In this work, the parameter $K$ is set to 4.

In general, the key to the proposed contrastive loss is to utilize diverse negative samples to push the enhanced image away from the degradation domain. From Figure \ref{fig5}, the negative sample generated by Eq. (\ref{eq6}) contains several degradation properties as follows:

\begin{itemize}
\item \textbf{Adverse illumination conditions}: The frequency-domain interpolation with dynamic parameter ${\lambda _{amp}}$ can control the illumination conditions of the frequency-domain interpolated images ${\rm{iFFT}}\left( {{{\mathcal P}_x}{\rm{, }}{{\mathcal A}_m}} \right)$ and ${\rm{iFFT}}\left( {{{\mathcal P}_y}{\rm{, }}{{\mathcal A}_m}} \right)$.
\item \textbf{Diverse noise and artifacts}: From Figure \ref{fig5}, there are also various noise and artifacts in the frequency-domain interpolated images, which is caused by the fact that amplitude and phase components are not independent (see Section \ref{Visibility-Restoration} for details). These degradations are simultaneously integrated into negative samples by spatial-domain interpolation.
\end{itemize}

\section{Experiment}
\subsection{Experiment Settings}

\begin{table*}[!t]
\caption{Quantitative results of different methods on SDE-in, SDE-out, SDSD-in and SDSD-out datasets. The best and second best results are bolded and underlined, respectively. `E', `F' and `E+F' denote Event, Frame, and Event+Frame, respectively. More quantitative results can be seen in Appendix B.}
\label{tab1}
\resizebox{\linewidth}{!}{
\begin{tabular}{l|l|ccc|ccc|ccc|ccc}
\hline
\multirow{2}{*}{Modality} & \multicolumn{1}{c|}{\multirow{2}{*}{Method}} & \multicolumn{3}{c|}{SDE-in}    & \multicolumn{3}{c|}{SDE-out}    & \multicolumn{3}{c|}{SDSD-in}    & \multicolumn{3}{c}{SDSD-out}   \\
\cline{3-14} 
&\multicolumn{1}{c|}{}                        & \multicolumn{1}{c}{PSNR↑} & \multicolumn{1}{c}{SSIM↑} & \multicolumn{1}{c|}{LPIPS↓} & \multicolumn{1}{c}{PSNR↑} & \multicolumn{1}{c}{SSIM↑} & \multicolumn{1}{c|}{LPIPS↓} & \multicolumn{1}{c}{PSNR↑} & \multicolumn{1}{c}{SSIM↑} & \multicolumn{1}{c|}{LPIPS↓} & \multicolumn{1}{c}{PSNR↑} & \multicolumn{1}{c}{SSIM↑} & \multicolumn{1}{c}{LPIPS↓} \\
\midrule
E & E2VID+ (ECCV’20) \cite{bib22}       & 15.19 & 0.4769 & 0.3803 & 14.99 & 0.4285 & 0.3848 & 13.51 & 0.5205 & 0.3456 & 16.57 & 0.3997 & 0.3826 \\
F & FourLLIE (ACM MM'23) \cite{bib19} & 20.94 & 0.6214 & 0.2274 & 21.06 & 0.6481 & 0.2591 & 25.54 & 0.8697 & 0.0681  & 25.14 & 0.7904 & 0.0879 \\
F & Retinexformer (ICCV'23) \cite{bib14} & 21.98 & 0.7211 & 0.2015 & \underline{23.70} & 0.7019 & 0.2596 & 28.01 & 0.8411 & 0.1196 & \textbf{27.49} & 0.8147 & 0.1131 \\
F & Reti-diff (ICLR'25) \cite{bib32}     & \underline{22.63} & 0.7470 & 0.1406 & 21.13 & 0.5668 & 0.2674 & \textbf{29.18} & \underline{0.9192} & 0.0481 & 25.26 & 0.8079 & 0.1191 \\
F+E & eSL-Net (ECCV'20) \cite{bib35}       & 21.32 & 0.7030 & 0.2184 & 23.00 & 0.7130 & 0.2251 & 25.53 & 0.8775 & 0.0959 & 24.57 & 0.8002 & 0.1743 \\
F+E & LLVE-SEG (AAAI'23) \cite{bib4}       & 21.71 & 0.6927 & 0.1714 & 22.41 & 0.6861 & 0.1696 & 27.68 & 0.8898 & 0.0534 & 22.84 & 0.7201 & 0.1201 \\
F+E & ELIE (TMM'24) \cite{bib6}           & 19.99 & 0.6179 & 0.2283 & 20.59 & 0.6493 & 0.2098 & 27.58 & 0.8831 & 0.0731 & 21.88 & 0.7232 & 0.1393 \\
F+E & EvLight (CVPR'24) \cite{bib5}        & 22.44 & \underline{0.7697} & \underline{0.1335} & 23.21 & \underline{0.7505} & \underline{0.1296} & 28.52 & 0.9125 & \underline{0.0408} & 26.67 & \underline{0.8356} & \underline{0.0691} \\
F+E & Ours                   & \textbf{23.18} & \textbf{0.7738} & \textbf{0.1266} & \textbf{23.99} & \textbf{0.7595} & \textbf{0.1148} & \underline{28.63} & \textbf{0.9255} & \textbf{0.0351} & \underline{27.06} & \textbf{0.8496} & \textbf{0.0678} \\
\bottomrule
\end{tabular}
}
\end{table*}

\textbf{Implementation Details}. We implement the proposed method with Pytorch and optimize it using Adam with a learning rate of $1.5e-4$. We randomly crop $256 \times 256$ patches for training and adopt the vertical and horizontal flips to augment the training set. The proposed method is trained for 100 epochs using an NVIDIA 4090 GPU. The batch size is set to 8. The parameters ${\lambda _{p}}$ and ${\lambda _{c}}$ are set to 0.5 and 0.1, respectively. All encoders and decoders use convolution and deconvolution for downsampling and upsampling, respectively.

\textbf{Datasets}. The performance of the proposed network is evaluated on several datasets, including the SDE dataset \cite{bib5} and the SDSD dataset with events \cite{bib5}. Each dataset contains an indoor subset and an outdoor subset. For simplicity, these four subsets are called SDE-in, SDE-out, SDSD-in, and SDSD-out, respectively. The SDSD dataset with events is generated by Liang et al. \cite{bib5} using the event simulator v2e \cite{bib28} and the dynamic version of the SDSD dataset \cite{bib27}. According to \cite{bib5}, we know that it is different from the static version of SDSD dataset used in common LLIE methods (e.g. Retinexformer \cite{bib14} and SNR-Net \cite{bib15}). Therefore, we follow the experimental protocol of \cite{bib5} and also retrain some LLIE methods without directly citing their results.

\subsection{Qualitative Evaluation}

We compare the proposed EventLLIE with some classical and state-of-the-art methods, including E2VID+ \cite{bib22}, FourLLIE \cite{bib19},  ELIE \cite{bib6}, LLVE-SEG \cite{bib4}, eSL-Net \cite{bib35}, Retinexformer \cite{bib14}, EvLight \cite{bib5} and Reti-diff \cite{bib32}. Details of these methods are provided in \textbf{Appendix A}. Figures \ref{fig6} and \ref{fig7} show the enhanced results of several high-performance methods on SDE-in, SDE-out, SDSD-in and SDSD-out datasets. We can observe several phenomena. First, event-based LLIE can restore more complete structure than frame-based LLIE. Second, although Reti-diff \cite{bib32} is a diffusion model-based enhancer, it carries the risk of over-smoothing. The enhanced results generated by EvLight \cite{bib5} also suffer from unclear structure and noise. In general, our method obtains satisfactory visibility and complete structure. More enhanced images are shown in \textbf{Appendix A}.

\subsection{Quantitative Evaluation}

We adopt the Peak Signal-to-Noise Ratio (PSNR), Structural Similarity (SSIM) \cite{bib36} and Learned Perceptual Image Patch Similarity (LPIPS) \cite{bib37} to quantitatively illustrate the competitiveness of the proposed EventLLIE. These metrics are often used for LLIE task. As shown in Table \ref{tab1}, the proposed EventLLIE achieves the best performance overall on the four datasets, which indicates its superiority in terms of both visibility restoration and structure refinement. Although it achieves only suboptimal results in terms of the PSNR metric on the SDSD dataset, it obtains the best SSIM and LPIPS scores across all datasets. This fact demonstrates that it effectively leverages event information, and its enhanced images exhibit rich and realistic structure compared with existing methods. These improvements have benefited from the following two factors. First, the visibility restoration network fully considers the inherent physical properties of low-light degradations, which is conducive to restoring favorable illumination conditions. Second, the proposed fusion strategy with dynamic alignment enables the events to accurately compensate the structure information of enhanced images. More quantitative results can be seen in \textbf{Appendix B}.

\begin{table}
  \caption{Quantitative results of ablation study on SDE-in.}
  \label{tab2}
   \scalebox{0.95}{
  \begin{tabular}{c|ccc}
    \toprule
    Method                      & PSNR↑  & SSIM↑   & LPIPS↓  \\
    \midrule
    
    w/o visibility restoration & 21.04 & 0.7313  & 0.1531 \\
    w/o structure refinement   & 21.10 & 0.6881  & 0.1717 \\
    w/o APE                    & 22.18 & 0.7565  & \underline{0.1424} \\
    w/o FDA                    & 21.54 & 0.7573  & 0.1460 \\
    w/o Eq. (\ref{eq6})        & 22.23 & 0.7633  & 0.1451 \\
    w/o contrastive loss       & \underline{22.34} & \underline{0.7667}  & 0.1450 \\
    Ours                       & \textbf{23.18}  & \textbf{0.7738} & \textbf{0.1266} \\
  \bottomrule
\end{tabular}
}
\end{table}

\begin{table*}
  \caption{Efficiency analysis for different methods.}
  \label{tab3}
  \resizebox{\linewidth}{!}{
  \begin{tabular}{c|ccccccccc}
    \toprule 
    Method     &  E2VID+ \cite{bib22} & FourLLIE \cite{bib19} & ELIE \cite{bib6} & LLVE-SEG \cite{bib4} & eSL-Net \cite{bib35} & EvLight \cite{bib5} & Retinexformer \cite{bib14} & Reti-diff \cite{bib32} & Ours  \\
    \midrule
    Param. (M)↓ &  55.98 & \textbf{0.12} & 33.36    & 47.06   & \underline{0.56}    & 22.73  & 1.61   & 26.11   & 15.21 \\
    FLOPs (G)↓ &  \underline{10.71}  &  \textbf{5.10} & 880.64 & 89.42 & 1121.88 & 361.80 & 31.14 & 313.10 & 203.57\\
     \bottomrule
\end{tabular}
}
\end{table*}

\subsection{Ablation Study}

The proposed EventLLIE decouples the LLIE task into two subtasks (i.e., visibility restoration and structure refinement). To verify the effectiveness of the proposed decoupling framework and modules, we conduct ablation experiments with six configurations on SDE-in dataset. \textbf{1)} "w/o visibility restoration" removes the visibility restoration network. \textbf{2)} "w/o structure refinement" removes the structure refinement network. \textbf{3)} "w/o APE" adopts a common Fourier module \cite{bib18} to replace the proposed amplitude-phase entanglement (APE) module. \textbf{4)} "w/o FDA" removes the proposed fusion strategy with dynamic alignment (FDA) and employs a simple concat operation to fuse the multiple features. \textbf{5)} "w/o Eq. (\ref{eq6})" means that the spatial-domain interpolation is abandoned in contrastive loss and only the frequency-domain interpolation of Eq. (\ref{eq5}) is adopted to generate negative samples, i.e., we take $\frac{K}{2}$ interpolated image ${\rm{iFFT}}\left( {{{\mathcal P}_x}{\rm{, }}{{\mathcal A}_m}} \right)$ and $\frac{K}{2}$ interpolated image ${\rm{iFFT}}\left( {{{\mathcal P}_y}{\rm{, }}{{\mathcal A}_m}} \right)$ as negative samples. \textbf{6)} "w/o contrastive loss" removes the proposed contrastive loss. The experimental results are reported in Table \ref{tab2}. We can find that each of the proposed strategies is effective. It is noteworthy that the performance of the fifth configuration is lower than that of the sixth configuration. One possible explanation is that the frequency-domain interpolated image ${\rm{iFFT}}\left( {{{\mathcal P}_y}{\rm{, }}{{\mathcal A}_m}} \right)$ may exhibit high similarity with the ground truth $y$, which prevents it from working effectively as a negative sample and consequently misguides the model's convergence. The visualization results of the ablation study are shown in \textbf{Appendix C}.

\subsection{Efficiency analysis}
Computational efficiency is one of the key factors to be considered for the LLIE task. We therefore report the number of parameters (Param.) and the floating point operations (FLOPs) in Table \ref{tab3}. In this work, the FLOPs is recorded at a resolution of $256 \times 256$. It can be found that the proposed EventLLIE does not increase computational overhead excessively compared to existing methods. As demonstrated in Tables \ref{tab1} and \ref{tab3}, the proposed EventLLIE presents promising superiority in terms of both performance and efficiency.

\subsection{Influence of parameter ${\lambda _{amp}}$}
\begin{figure}
    \centering
    \subfigure
    {
        \includegraphics[width=0.233\textwidth]{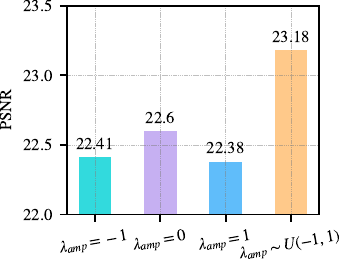}
    }
    \hspace{-3mm}
    \subfigure
    {
        \includegraphics[width=0.233\textwidth]{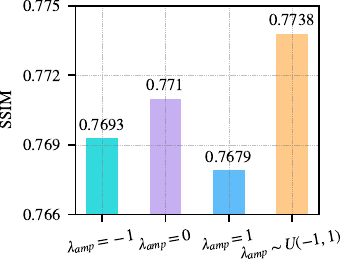}
    }
    \caption{Comparison of different parameters ${\lambda _{amp}}$.}
    \label{fig8}
\end{figure}

To obtain the negative sample with diverse degradations, the parameter ${\lambda _{amp}}$ is sampled dynamically from the uniform distribution $U\left( { - 1,1} \right)$. To examine the effectiveness of this strategy, we compare it to the fixed parameter ${\lambda _{amp}}$. Figure \ref{fig8} shows the competitiveness of the dynamic parameter. Furthermore, there are two interesting phenomena here. First, when ${\lambda _{amp}}$ is set to $1$ and $-1$, the power of contrastive loss is severely suppressed. Second, it achieves sub-optimal performance when ${\lambda _{amp}}$ is set to $0$. One possible reason is that negative samples with ${\lambda _{amp}}$ equal to $0$ have a stronger strictness than others, which better encourages the model to generate high-quality results.

\subsection{More Analysis}
The key components of this paper are the amplitude-phase entanglement module and the fusion strategy with dynamic alignment. To further examine their superiority, they will be compared with similar works.

\textbf{Discussion of Fourier Module}. To further illustrate the superiority of the proposed amplitude-phase entanglement module, we compare it with several Fourier modules employed in popular works \cite{bib38,bib39}. As shown in Table \ref{tab4}, we can find that the proposed amplitude-phase entanglement module has stronger competitiveness compared with existing Fourier modules.

\textbf{Discussion of Fusion Strategy}. We also explore the performance of different fusion strategies. Considering that three types of features need to be fused here, we employ several classical and applicable fusion strategies to replace the proposed fusion strategy with dynamic alignment, including SKC \cite{bib40} and SNR-Fusion \cite{bib5}. It is worth mentioning that the SNR-Fusion \cite{bib5} is customized to fuse event and image features. Table \ref{tab4} presents the effectiveness of the proposed fusion strategy with dynamic alignment. Although SNR-Fusion \cite{bib5} achieves the best SSIM score, its other metrics are unsatisfactory. This fact indirectly indicates that the proposed dynamic alignment accurately utilizes events as complementary information to assist the LLIE task.

\begin{table}
  \caption{Quantitative results of different Fourier modules and fusion modules on SDE-in.}
  \label{tab4}
  \scalebox{0.88}{
  \begin{tabular}{c|c|ccc}
    \toprule    
    Fourier module            & Fusion module       & PSNR↑  & SSIM↑   & LPIPS↓  \\
    \midrule
    FSAS \cite{bib38}          
& \multirow{3}{*}{FDA (Ours)}&    22.28 & 0.7650 & 0.1482       \\
                                    SFFB \cite{bib39}       
& &      \underline{22.44} & \underline{0.7702}& \underline{0.1402}        \\
                                    APE (Ours)& & \textbf{23.18} & \textbf{0.7738} & \textbf{0.1266} \\
    \midrule
    \multirow{3}{*}{APE (Ours)}& SKC \cite{bib40}                    & \underline{22.42} & 0.7602 & \underline{0.1347} \\
                                    & SNR-Fusion \cite{bib5}                   & 21.96 & \textbf{0.7760}& 0.1452 \\
                                    & FDA (Ours)& \textbf{23.18} & \underline{0.7738} & \textbf{0.1266} \\  
    \bottomrule
\end{tabular}
}
\end{table}

\section{Conclusion}

In this work, we propose an event-based LLIE that decouples the enhancement pipeline into two stages: visibility restoration and structure refinement. In the first stage, we employ Fourier prior to design a visibility restoration network with amplitude-phase entanglement. Compared with the traditional Fourier prior, we rethink the relationship between amplitude and phase components. In the second stage, we observe a spatial mismatch issue between events and image's structure, which is caused by their temporal resolution discrepancy. To this end, we propose a dynamic alignment strategy to fuse two complementary modalities, aiming to compensate for the structure information of enhanced images. Besides, a contrastive loss based on the spatial-frequency interpolation is developed to further encourage the enhancer to learn degradation-invariant feature representations. In the future, we will explore interpretable approaches to implement event-based LLIE task.

\begin{acks}
This work is supported by the National Natural Science Foundation of China (Grant Nos.U20A20227, 62076208, 62076207), Chongqing Talent Plan Project (Grant No.CQYC20210302257), Fundamental Research Funds for the Central Universities (No.SWU-XDZD22009), Chongqing Higher Education Teaching Reform Research Project (Grant No.211005) and Graduate Research Innovation Project of Southwest University (Grant No.SWUB24075)
\end{acks}


\bibliographystyle{ACM-Reference-Format}
\bibliography{sample-base}

\end{document}